\documentclass{article}

\PassOptionsToPackage{numbers, compress}{natbib}

 \usepackage[preprint]{nips_2018}

\usepackage[T1]{fontenc}    
\usepackage{url}            
\usepackage{booktabs}       
\usepackage{amsfonts}       
\usepackage{nicefrac}       
\usepackage{microtype}      
\usepackage{tabularx}
\usepackage{algorithm}
\usepackage{algorithmic}
\usepackage{epsfig}
\usepackage{amsmath}
\usepackage{multirow}
\usepackage{subfigure}

\title{Graph Learning-Convolutional Networks}

\author{
  Bo Jiang, Ziyan Zhang, Doudou Lin, Jin Tang\\
  School of Computer Science and Technology \\
  Anhui University\\
  Hefei, China \\
  \texttt{jiangbo@ahu.edu.cn} \\
}

\begin{document}

\maketitle

\begin{abstract}
Recently, graph Convolutional Neural Networks (graph CNNs) have been widely used for graph data representation and semi-supervised learning tasks.
However, existing graph CNNs generally use a fixed graph which may be not optimal for semi-supervised learning tasks.
In this paper, we propose a novel Graph Learning-Convolutional Network (GLCN) for graph data representation and semi-supervised learning.
The aim of GLCN is to learn an optimal graph structure that best serves graph CNNs for semi-supervised learning by integrating both \textbf{graph learning} and \textbf{graph convolution} together in a unified network architecture.
The main advantage is that in GLCN, both given labels and the estimated labels are incorporated and thus can provide useful `weakly' supervised information to refine (or learn) the graph construction and also to facilitate the graph convolution operation in GLCN for unknown label estimation.
Experimental results on seven benchmarks demonstrate that GLCN significantly outperforms state-of-the-art traditional fixed structure based graph CNNs.

\end{abstract}

\section{Introduction}

Deep neural networks have been widely used in computer vision and pattern recognition.
In particular, Convolutional Neural Networks (CNNs)
have been successfully applied in many problems, such as object detection and recognition, in which
the underlying data generally have grid-like structure.
However, in many real applications, data usually have irregular structure forms which are generally represented as structured graphs.
Traditional CNNs generally fail to address graph structure data.

Recently, many methods have been proposed to generalize the convolution operator on arbitrary graphs \cite{duvenaud2015convolutional,atwood2016diffusion,monti2017geometric,kipf2016semi,adaptive_GCN,velickovic2017graph}.
Overall, these methods can be categorized into spatial convolution and spectral convolution methods.
For spatial methods,
they generally define graph convolution operation directly  by defining an operator on node groups of neighbors.
For example,
Duvenaud et al. \cite{duvenaud2015convolutional}
propose a convolutional neural network that operates directly on graphs and
 provide an end-to-end feature learning for graph data.
Atwood and Towsley \cite{atwood2016diffusion} propose Diffusion-Convolutional Neural Networks (DCNNs) by
employing a graph diffusion process to incorporate the contextual information of node in graph node  classification.
Monti et al. \cite{monti2017geometric} present mixture model CNNs (MoNet) and provide a unified generalization of CNN architectures on graphs.
Velickovic et al. \cite{velickovic2017graph} present Graph Attention Networks (GAT) for semi-supervised learning by designing an attention layer. 
For spectral methods, they generally  define graph convolution operation based on spectral representation of graphs.
For example,
Bruna et al. \cite{bruna2014spectral} propose to define graph convolution in the Fourier domain based on eigen-decomposition of graph Laplacian matrix.
Defferrard et al. \cite{defferrard2016convolutional} propose to approximate the spectral filters based on Chebyshev expansion of graph
Laplacian to avoid the high computational complexity of eigen-decomposition.
Recently, Kipf et al. \cite{kipf2016semi}  propose a more simplified Graph Convolutional Network (GCN) for semi-supervised learning by employing a first-order approximation of spectral filters.

The above graph CNNs have been widely used for supervised or semi-supervised learning tasks.
In this paper, we focus on semi-supervised learning.
One important aspect of graph CNNs is the graph structure representation of data.
In general, the data we feed to graph CNNs either has a known intrinsic graph structure, such as social networks, or we construct a human established graph for it, such as $k$-nearest neighbor graph with Gaussian kernel.
However, it is difficult to evaluate whether the graphs obtained from domain knowledge (e.g., social network) or established by human are optimal for semi-supervised learning in graph CNNs.
Henaff et al. \cite{henaff2015deep} propose to learn a supervised graph with a fully connected network.
However, the learned graph is obtained from a separate network which is also not guaranteed to best serve the graph CNNs.
Li et al. \cite{velickovic2017graph} propose optimal graph CNNs, in which the graph is learned adaptively by using a traditional distance metric learning.
However, it use an approximate algorithm to estimate graph Laplacian which may lead to weak local optimal solution.

\begin{figure*}[!htbp]
\centering
\centering
\includegraphics[width=0.975\textwidth]{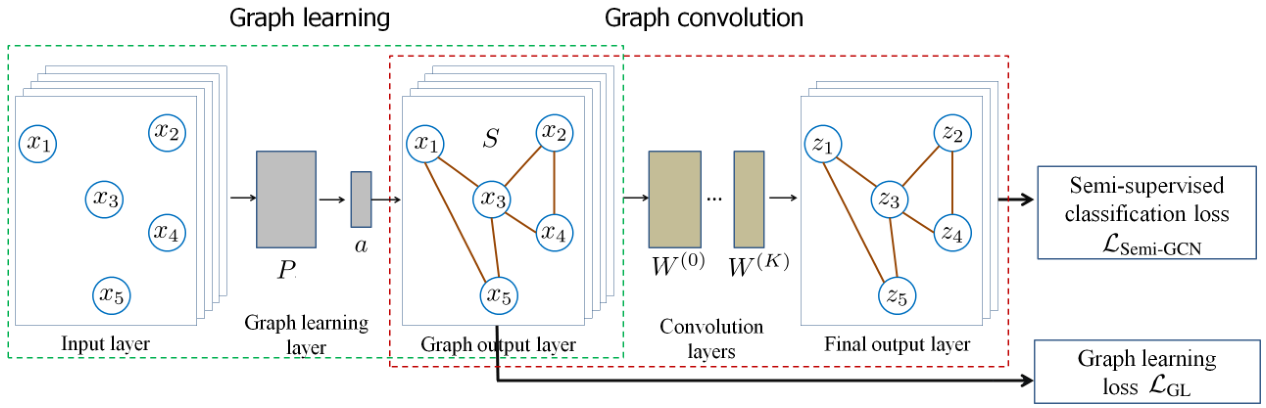}
  \caption{Architecture of the proposed GLCN network for semi-supervised learning. }\label{fig::lambda}
\end{figure*}
In this paper, we propose a novel Graph Learning-Convolutional Network (GLCN) for semi-supervised learning problem.
The main idea of GLCN is to learn an optimal graph representation that best serves graph CNNs for semi-supervised learning by integrating both \emph{graph learning} and \emph{graph convolution} simultaneously in a unified network architecture. 
The main advantages of the proposed GLCN for semi-supervised learning are summarized as follows.
\begin{itemize}
  \item In GLCN, both given labels and the estimated labels are incorporated and thus can provide useful `weakly' supervised information to refine (or learn) the graph construction and to facilitate the graph convolution operation in graph CNN for unknown label estimation.
  \item GLCN can be trained via a single optimization manner, which can thus be implemented simply.
\end{itemize}
%
To the best of our knowledge, this is the first attempt to build a \emph{unified graph learning-convolutional network architecture} for semi-supervised learning. 
Experimental results on seven benchmarks demonstrate that GLCN significantly outperforms state-of-the-art graph CNNs on
semi-supervised learning tasks.

\section{Related Work}

Recently, graph convolutional network (GCN) \cite{defferrard2016convolutional,kipf2016semi} has been commonly used
to address structured graph data.
In this section, we briefly review GCN based semi-supervised learning proposed in ~\cite{kipf2016semi}.
%

Let $X=(x_1,x_2,\cdots x_n)\in \mathbb{R}^{n\times p}$ be the collection of $n$ data vectors in $p$ dimension.
Let $G(X, A)$ be the graph representation of $X$ with $A \in \mathbb{R}^{n\times n}$ encoding the pairwise relationship (such as similarities, neighbors) among data $X$.
%
GCN contains one input layer, several propagation (hidden) layers and one final perceptron layer~\cite{kipf2016semi}.
Given an input $X^{(0)} = X$ and graph  $A$, GCN conducts the following layer-wise propagation in hidden layers as ~\cite{kipf2016semi}, 
%
\begin{align}\label{EQ:layer_gcn}
X^{(k+1)} = \sigma({D}^{-1/2}A{D}^{-1/2}X^{(k)}W^{(k)})
\end{align}
%
where $k=0,1,\cdots K-1$. ${D}=\textrm{diag}(d_1,d_2\cdots d_n)$ is a diagonal matrix with
$d_{i}=\sum^n_{j=1} {A}_{ij}$ and $W^{(k)}\in \mathbb{R}^{d_{k}\times d_{k+1}}, d_0 = p$ is a layer-specific weight matrix needing to be trained.
$\sigma(\cdot)$ denotes an activation function, such as $\mathrm{ReLU}(\cdot) = \max(0,\cdot)$, and $X^{(k+1)} \in \mathbb{R}^{n\times d_{k+1}}$ denotes the output of activations in the $(k+1)$-th layer.



For semi-supervised node classification, GCN defines the final perceptron layer as
\begin{equation}\label{EQ:final_gcn}
Z = \mathrm{softmax} ({D}^{-1/2}A{D}^{-1/2}X^{(K)} W^{(K)} )
\end{equation}
where $W^{(K)}\in \mathbb{R}^{d_{K}\times c}$ and $c$ denotes the number of classes.
The final output $Z\in \mathbb{R}^{n\times c}$ denotes the label prediction for all data $X$ in which each row $Z_i$ denotes the label prediction for the $i$-th node.
The optimal weight matrices $\mathcal{W} = \{W^{(0)}, W^{(1)},\cdots W^{(K)}\}$ are trained  by minimizing the following cross-entropy loss function over all the labeled nodes $L$, i.e.,
 \begin{equation}\label{EQ:GCN_semi}
\mathcal{L}_{\textrm{Semi-GCN}} = -\sum\nolimits_{i\in L} \sum^c\nolimits_{j=1} Y_{ij}\mathrm{ln} Z_{ij}
 \end{equation}
where ${L}$ indicates the set of labeled nodes and $Y_{i\cdot}, i\in L$ denotes the corresponding label indication for the $i$-th labeled node.\\
\noindent \textbf{Remark.} By using Eq.(\ref{EQ:layer_gcn}) and Eq.(\ref{EQ:final_gcn}), GCN indeed provides a kind of nonlinear function $Z = \mathcal{F}_{\textrm{GCN}}(X,A,Y;\mathcal{W})$ to predict the labels for graph nodes.

%
%

\section{Graph Learning-Convolutional Network}

One core aspect of GCN is the graph representation $G(X,A)$ of data $X$.
In some applications,
 the graph structure of data are available from domain knowledge, such as chemical molecules, social networks etc.
 In this case, one can use the existing graph directly for GCN based semi-supervised learning.
In many other applications, the graph data are not available. One popular way is to construct a human established graph (e.g., $k$-nearest neighbor graph) for GCN.
However, the graphs obtained from domain knowledge or estimated by human are generally independent of GCN (semi-supervised) learning process
and thus are not guaranteed to best serve GCN learning.
Also, the human established graphs are usually sensitive to the local noise and outliers.
%
%
To overcome these problems,
 we propose a novel Graph Learning-Convolution Network (GLCN) which
  integrates both graph learning and graph convolution simultaneously in a unified network architecture and thus can learn an adaptive (or optimal) graph representation for GCN  learning.
As shown in Figure 1, GLCN contains one graph learning layer, several convolution layers and one final perceptron layer. In
the following, we explain them in detail.
\begin{figure*}[htpb]
\centering
\centering
\includegraphics[width=0.68\textwidth]{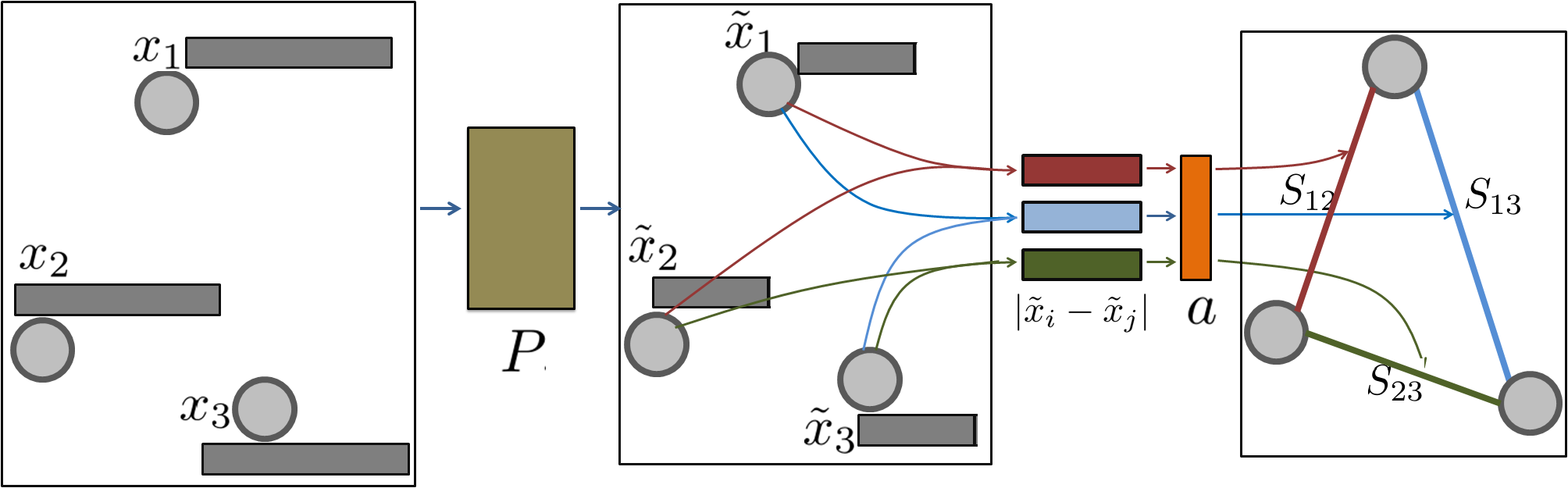}
  \caption{Architecture of the proposed graph learning  architecture in GLCN.}\label{fig::lambda}
\end{figure*}
\subsection{Graph learning architecture}

Given an input $X=(x_1,x_2\cdots x_n)\in \mathbb{R}^{p\times n}$, 
we aim to seek a nonnegative function $S_{ij} = g(x_i,x_j)$ that represents the pairwise relationship between data $x_i$ and $x_j$.
We implement $g(x_i,x_j)$ via a single-layer neural network, which is parameterized by a weight vector
$ {a}=(a_1,a_2,\cdots a_p)^T\in \mathbb{R}^{p\times 1}$. Formally, we learn a graph $S$ as
%
\begin{equation}\label{}
S_{ij} = g(x_i,x_j) = \frac{\exp(\textrm{ReLU}( {a}^T| {x}_i -  {x}_j|))}{\sum^n_{j=1}\exp(\textrm{ReLU}( {a}^T| {x}_i -  {x}_j|))}
\end{equation}
where $\textrm{ReLU}(\cdot)=\max(0,\cdot)$ is an activation function, which guarantees the nonnegativity of $S_{ij}$.
 The role of the above softmax operation on each row of $S$ is to guarantee that the learned graph $S$ can satisfy the following property,
\begin{equation}\label{EQ:constraint}
\sum\nolimits^n_{j=1} S_{ij} =1, S_{ij}\geq 0
\end{equation}
We optimize the optimal weight vector $a$ by minimizing the following loss function,
%
\begin{equation}\label{EQ:loss_GM}
\mathcal{L}_{\textrm{GL}} = \sum\nolimits^n_{i,j=1}\|x_i - x_j\|^2_2 S_{ij} +\gamma \|S\|^2_F
\end{equation}
That is,
larger distance $\|x_i - x_j\|_2$ between data point $x_i$ and $x_j$ encourages a smaller value $S_{ij}$.
The second term is used to control the sparsity of learned graph $S$ because of simplex property of $S$ (Eq.(\ref{EQ:constraint})), as discussed in \cite{nie2014clustering}. 

\noindent\textbf{Remark.}
Minimizing the above loss $\mathcal{L}_{\textrm{GL}}$ independently may lead to trivial solution, i.e., $a=(0,0\cdots 0)$.
We use it as a regularized term in our final loss function, as shown in Eq.(\ref{EQ:final_loss}) in \S3.2.

For some problems, when an initial graph $A$ is available, we can incorporate it in our graph learning as
\begin{equation}\label{}
S_{ij} = g(x_i,x_j) = \frac{A_{ij}\exp(\textrm{ReLU}( {a}^T| {x}_i -  {x}_j|))}{\sum^n_{j=1}A_{ij}\exp(\textrm{ReLU}( {a}^T| {x}_i -  {x}_j|))}
\end{equation}
We can also  incorporate the information of $A$ by considering a regularized term in the learning loss function as
%
\begin{equation}\label{EQ:loss_GM}
\mathcal{L}_{\textrm{GL}} = \sum\nolimits^n_{i,j=1}\|x_i - x_j\|^2_2 S_{ij} + \gamma \|S\|^2_F + \beta \|S - A\|^2_F
\end{equation}
%


On the other hand, when the dimension $p$ of the input data $X$ is large, the above computation of $g(x_i,x_j)$ may be less effective due to the long weight vector $a$ needing to be trained.
Also, the computation of Euclidean distances $\|x_i-x_j\|_2$ between data pairs in loss function $\mathcal{L}_{\textrm{GL}}$ is complex for large dimension $p$. 
To solve this problem, we propose to conduct our graph learning in a low-dimensional subspace.
We implement this via a single-layer low-dimensional embedding network, parameterized by a projection matrix $P\in \mathbb{R}^{p\times d}, d< p$.
In particular, we conduct our final graph learning  as follows,
\begin{align}\label{graph_learning_final}
& \tilde{x}_i = x_i P,\ \ \text{for}\ \ i=1,2\cdots n  \\
& S_{ij} = g(\tilde{x}_i,\tilde{x}_j) =  \frac{A_{ij}\exp(\textrm{ReLU}( {a}^T| \tilde{x}_i -  \tilde{x}_j|))}{\sum^n_{j=1}A_{ij}\exp(\textrm{ReLU}( {a}^T| \tilde{x}_i -  \tilde{x}_j|))}
\end{align}
where $A$ denotes an initial graph. If it is unavailable, we can set $A_{ij}=1$  in the above update rule.
The loss function becomes
\begin{equation}\label{EQ:loss_GM}
\mathcal{L}_{\textrm{GL}} = \sum\nolimits^n_{i,j=1}\|\tilde{x}_i - \tilde{x}_j\|^2_2 S_{ij}  + \gamma \|S\|^2_F
\end{equation}
The whole architecture of the proposed graph learning network is shown in Figure 2.

%
%

%

 \textbf{Remark.}
The proposed learned graph $S$ has a desired probability property (Eq.(\ref{EQ:constraint})), i.e.,
the optimal $S_{ij}$ can be regarded a probability that data $x_j$ is connected to $x_i$ as
a neighboring node. That is, the proposed graph learning (GL) architecture can establish the neighborhood structure of data automatically either based on data feature $X$ only or by further incorporating the prior initial graph $A$ with $X$.
The GL architecture
indeed provides a kind of nonlinear function $S=\mathcal{G}_{\textrm{GL}}(X,A;P,a)$ to
predict/compute the neighborhood probabilities between node pairs.

\subsection{GLCN architecture}

The proposed graph learning architecture is general and can be incorporated in any graph CNNs.
In this paper, we incorporate it into GCN~\cite{kipf2016semi} and propose a unified Graph Learning-Convolutional Network (GLCN) for semi-supervised learning problem. Figure 1 shows the overview of GLCN architecture.
The aim of GLCN is to learn an optimal graph representation for GCN network and integrates graph learning and convolution simultaneously to boost their respectively performance.

As shown in Figure 1, GLCN contains one graph learning layer, several graph convolution layers and one final perceptron layer.
The graph learning layer aims to provide an optimal adaptive graph representation $S$ for the following graph convolutional layers.
That is, in the convolutional layers, it conducts the layer-wise propagation rule based on the adaptive neighbor graph $S$ returned by graph learning layer, i.e., 
\begin{align}\label{EQ:layer}
X^{(k+1)} = \sigma({D}_s^{-1/2}S{D}_s^{-1/2}X^{(k)}W^{(k)})
\end{align}
%
where $k=0,1\cdots K-1$. ${D}_s=\textrm{diag}(d_1,d_2,\cdots d_n)$ is a diagonal matrix with diagonal element
$d_{i}=\sum^n_{j=1} {S}_{ij}$. $W^{(k)}\in \mathbb{R}^{d_{k}\times d_{k+1}}$ is a layer-specific trainable weight matrix for each convolution layer.
$\sigma(\cdot)$ denotes an activation function, such as $\mathrm{ReLU}(\cdot) = \max(0,\cdot)$, and $X^{(k+1)} \in \mathbb{R}^{n\times d_{k+1}}$ denotes the output of activations in the $k+1$-th layer.
Since the learned graph $S$ satisfies $\sum_j S_{ij}=1, S_{ij}\geq 0$, thus Eq.(\ref{EQ:layer}) can be simplified as
\begin{align}\label{EQ:layer}
X^{(k+1)} = \sigma(SX^{(k)}W^{(k)})
\end{align}
%


For semi-supervised classification task, we define the final perceptron layer as
\begin{equation}\label{EQ:final}
Z = \mathrm{softmax} (SX^{(K)} W^{(K)} )
\end{equation}
where $W^{(K)}\in \mathbb{R}^{d_{K}\times c}$ and $c$ denotes the number of classes.
The final output $Z\in \mathbb{R}^{n\times c}$ denotes the label prediction of GLCN network, in which each row $Z_i$ denotes the label prediction for the $i$-th node.
The whole network parameters $\Theta = \{P, a, W^{(0)}, \cdots W^{(K)}\}$ are jointly trained  by minimizing the following  loss function as 
 \begin{equation}\label{EQ:final_loss}
\mathcal{L}_{\textrm{Semi-GLCN}} =  \mathcal{L}_{\textrm{Semi-GCN}} +\lambda\mathcal{L}_{\textrm{GL}}
 \end{equation}
where $\mathcal{L}_{\textrm{GL}}$ and $\mathcal{L}_{\textrm{Semi-GCN}}$ are defined in Eq.(\ref{EQ:loss_GM}) and Eq.(\ref{EQ:GCN_semi}), respectively. Parameter $\lambda \geq0$ is a tradeoff parameter. 
It is noted that, when $\lambda=0$, the optimal graph $S$ is learned based on labeled data (i.e., cross-entropy loss) only which is also feasible in our GLCN.


\noindent \textbf{Demonstration and analysis. }
There are two main benefits of the proposed GLCN network:
\begin{itemize}
  \item In GLCN, both given labels $Y$ and the estimated labels $Z$ are incorporated and thus can provide useful `weakly' supervised information to refine the graph construction $S$ and thus to facilitate the graph convolution operation in GCN for unknown label estimation. That is, the graph learning and
      semi-supervised learning are conducted jointly in GLCN and thus can boost their respectively performance.
  \item GLCN is a unified network which can be trained via a single optimization manner and thus can be implemented simply.
\end{itemize}
Figure 3 shows the cross-entropy loss values over labeled node $L$ across different epochs.
One can note that, GLCN obtains obviously lower cross-entropy value than GCN at convergence, which clearly demonstrates the higher predictive accuracy of GLCN model.
Also, the convergence speed of GLCN is just slightly slower than GCN, indicating the efficiency of GLCN.
Figure 4 demonstrates 2D t-SNE \cite{Geoffrey2017Visualizing} visualizations of the feature map output by the first convolutional layer of GCN~\cite{kipf2016semi} and GLCN, respectively. Different classes are marked by different colors. One can note that, the data of different classes are distributed more clearly  and compactly in our GLCN  representation, which demonstrates the desired discriminative ability of GLCN on conducting graph node representation and thus semi-supervised classification tasks.

%
\begin{figure}[htpb]
\centering
\centering
\includegraphics[width=0.5\textwidth]{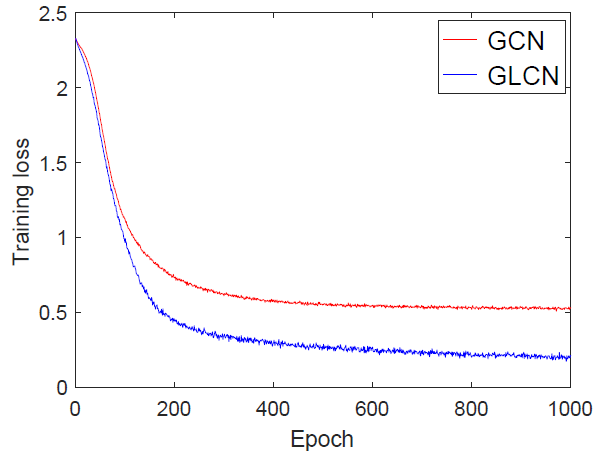}
  \caption{Demonstration of cross-entropy loss values across different epochs on MNIST dataset.}\label{fig::lambda}
\end{figure}
\begin{figure*}[!htpb]
\centering
\centering
\includegraphics[width=0.9\textwidth]{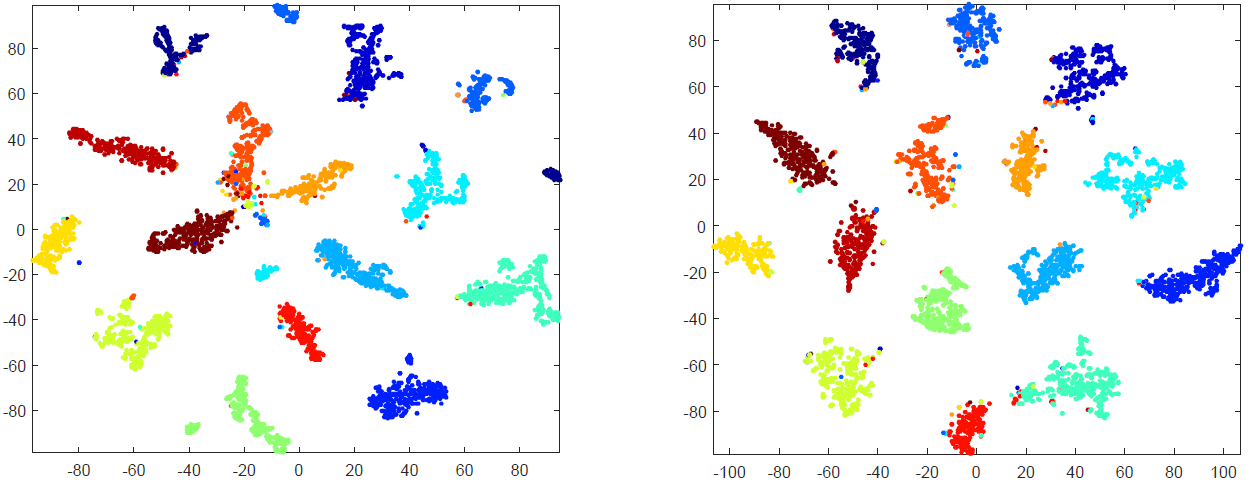}
  \caption{2D t-SNE \cite{Geoffrey2017Visualizing} visualizations of the feature map output by the first convolutional layer of GCN~\cite{kipf2016semi} and GLCN  respectively on Scene15 dataset. Different classes are marked by different colors. One can note that, the data of different classes are distributed more clearly  and compactly in our GLCN convolutional layer feature representation.}\label{fig::lambda}
\end{figure*}

\section{Experiments}

\subsection{Datasets}
To verify the effectiveness and benefit of the proposed GLCN on semi-supervised learning tasks, we test it on seven benchmark datasets, including three  standard citation network benchmark datasets (Citeseer, Cora and Pubmed ~\cite{sen2008collective}) and four image datasets (CIFAR10~\cite{krizhevsky2009learning}, SVHN~\cite{netzer2011reading}, MNIST and Scene 15 ~\cite{jiang2013label}).
The details of these datasets and their usages in our experiments are introduced below. 

\noindent\textbf{Citeseer.}
This data contains 3327 nodes and 4732 edges.
Each node corresponds to a document and edge to citation relationship between
documents.
The nodes of this network are classified into 6 classes and
each node has been represented by a 3703 dimension feature descriptor.
\\
\noindent\textbf{Cora.}
This data contains 2708 nodes and 5429 edges.
Nodes correspond to documents and edges to citations between documents.
Each node has a 1433 dimension feature descriptor and all the nodes are classified  into 6 classes. 
\\
\noindent\textbf{Pubmed.}
This data contains 19717 nodes and 44338 edges.
Each node is represented by a 500 dimension feature descriptor and
all the nodes are classified into 3 classes.
%
\\
\noindent\textbf{CIFAR10.}
This dataset contains  50000 natural images which are falling into 10 classes ~\cite{krizhevsky2009learning}.
Each image in this dataset is a $32\times 32$ RGB color image.
In our experiments, we select 1000 images for each class and use 10000 images in all for
our evaluation.
We have not use all of images for our evaluation because large storage and high computational complexity are required for graph convolution operation in our GLCN and other comparing GCN based methods.
For each image, we extract a  CNN feature descriptor for it.
\\
\noindent\textbf{SVHN.}
It contains 73257 training and 26032 test images ~\cite{netzer2011reading}.
Each image is a $32\times 32$ RGB image which contains  multiple number of  digits and
the task is to recognize the digit in the image center.
Similar to CIFAR 10 dataset, in our experiments, we select 1000 images for each class and
obtain 10000 images in all for our evaluation.
For each image, we extract a CNN feature descriptor for it.
\\
\noindent\textbf{MNIST.}
This dataset consists of images of hand-written digits from `0' to `9'.
Each image is centered on a $28\times 28$ grid.
Here, we randomly select 1000 images from each digit class and obtain 10000 images in all for our evaluation.
Similar to other related works, we use gray value directly and convert it to a 784 dimension vector.
\\
\noindent\textbf{Scene15.}
It consists of 4485 scene images with 15 different categories~\cite{jiang2013label}.
For each image, we use the feature descriptor provided by work~\cite{jiang2013label}.
\begin{table}[!htp]
\centering
\caption{Comparison results of semi-supervised learning on dataset Citeseer, Cora and Pubmed.}
\centering
\begin{tabular}{c|c|c|c}
  \hline
  \hline
  Methond & Citeseer & Cora & Pummed\\
  \hline
  ManiReg~\cite{belkin2006manifold} & 60.1\% & 59.5\% & 70.7\%\\
  LP~\cite{zhu2003semi} & 59.6\% & 59.0\% & 71.1\%\\
  DeepWalk~\cite{perozzi2014deepwalk} & 43.2\% &67.2\% & 65.3\%\\
  GCN~\cite{kipf2016semi} & 68.9\% & 82.9\% & 77.9\%\\
  GAT ~\cite{velickovic2017graph} & 71.0\% & 83.2\% & 78.0\%\\
  \hline
  GLCN & \textbf{72.0\%} & \textbf{85.5\%} & \textbf{78.3\%}\\
  \hline
  \hline
\end{tabular}
\end{table}
%
\begin{table*}[!htp]\footnotesize
\centering
\caption{Comparison results on dataset SVHN, CIFAR, MNIST and Scene15}
\centering
\begin{tabular}{c||c|c|c||c|c|c}
  \hline
  \hline
  Dataset& \multicolumn{ 3}{c||}{SVHN} & \multicolumn{ 3}{c}{CIFAR}\\
  \hline
  No. of label & 1000 & 2000 & 3000 & 1000 & 2000 & 3000 \\
  \hline
  ManiReg~\cite{belkin2006manifold} &69.44$\pm$0.69 &72.73$\pm$0.44 &74.63$\pm$0.45 &52.30$\pm$0.66 &57.08$\pm$0.80 &59.69$\pm$0.71 \\
  LP~\cite{zhu2003semi} & 69.68$\pm$0.84 & 70.35$\pm$1.73 & 69.47$\pm$2.96 & 57.52$\pm$0.67 & 59.22$\pm$0.67 & 60.38$\pm$0.51 \\
  DeepWalk~\cite{perozzi2014deepwalk} &74.64$\pm$0.23 &76.21$\pm$0.23 &77.04$\pm$0.42 &56.16$\pm$0.54 &59.73$\pm$0.35 &61.26$\pm$0.32 \\
  GCN~\cite{kipf2016semi} & 71.33$\pm$1.48 & 73.43$\pm$0.46 & 73.63$\pm$0.52 & 60.43$\pm$0.56 & 60.91$\pm$0.50 & 60.99$\pm$0.49  \\
  GAT ~\cite{velickovic2017graph} & 73.87$\pm$0.32 & 74.85$\pm$0.55 & 75.17$\pm$0.43 & 63.25$\pm$0.50 & 65.55$\pm$0.58 & 66.56$\pm$0.58\\
  \hline
  GLCN & \textbf{79.14$\pm$0.38} & \textbf{80.68$\pm$0.22} & \textbf{81.43$\pm$0.34} & \textbf{66.67$\pm$0.24} & \textbf{69.33$\pm$0.54} & \textbf{70.39$\pm$0.54}  \\
  \hline
  \hline
  Dataset & \multicolumn{ 3}{c||}{MNIST} & \multicolumn{ 3}{c}{Scene15}\\
  \hline
  No. of label  & 1000 & 2000 & 3000 & 500 & 750 & 1000\\
  \hline
  ManiReg~\cite{belkin2006manifold}  &92.74$\pm$0.33 &93.96$\pm$0.23 &94.62$\pm$0.22 &81.29$\pm$3.35 &86.45$\pm$1.91 &89.86$\pm$0.71 \\
  LP~\cite{zhu2003semi} &79.28$\pm$0.91 & 81.91$\pm$0.82 & 83.45$\pm$0.53 & 89.40$\pm$4.74 & 92.12$\pm$2.87 & 92.98$\pm$2.45   \\
  DeepWalk~\cite{perozzi2014deepwalk}  & \textbf{94.55$\pm$0.27} & 95.04$\pm$0.28 & 95.34$\pm$0.26 & 95.64$\pm$0.24 & 96.01$\pm$0.24 & 96.53$\pm$0.37  \\
  GCN~\cite{kipf2016semi} & 90.59$\pm$0.26 & 90.91$\pm$0.19 & 91.01$\pm$0.23 & 91.42$\pm$2.07 & 94.41$\pm$0.92 & 95.44$\pm$0.89   \\
  GAT ~\cite{velickovic2017graph}  & 92.11$\pm$0.35 & 92.64$\pm$0.28 & 92.81$\pm$0.29 & 93.98$\pm$0.75 & 94.64$\pm$0.41 & 95.03$\pm$0.46\\
  \hline
  GLCN & 94.28$\pm$0.28 & \textbf{95.09$\pm$0.17} & \textbf{95.46$\pm$0.20}& \textbf{96.19$\pm$0.38} & \textbf{96.71$\pm$0.40} & \textbf{96.67$\pm$0.37}  \\
  \hline
  \hline
\end{tabular}
\end{table*}
\subsection{Experimental setting}

For Cora, Citeseer and Pubmed datasets,
we follow the experimental setup of previous works ~\cite{kipf2016semi,velickovic2017graph}.
That is, for each class, we select 20 nodes as label data and evaluate the performance of label prediction on 1000 test nodes.
In addition, we use 300 additional nodes for validation, which is same as setting in ~\cite{kipf2016semi,velickovic2017graph}.
Note that, for graph based semi-supervised setting, we use all of the nodes in our network training.
For image dataset CIFAR10~\cite{krizhevsky2009learning}, SVHN~\cite{netzer2011reading} and MNIST,
we randomly select 1000, 2000 and 3000 images as labeled samples and the
remaining data are used as unlabeled samples.
For unlabeled samples, we select 1000 images for validation purpose and
use the remaining 8000, 7000 and 6000 images as test samples, respectively.
All the reported results are averaged over 10 runs with different groups of training, validation and testing data splits.
For image dataset Scene15~\cite{jiang2013label},
we randomly select 500, 750 and 1000 images as label data and
use 500 images for validation, respectively.
The remaining samples are used as the unlabeled test samples.
The reported results are averaged over 10 runs  with different groups of training, validation and testing data splits.

Similar to \cite{kipf2016semi}, we set the number of convolution layers in our GLCN to 2.
The number of units in each hidden layer is set to 70.
We provide additional experiments on different number of units and convolutional layers in \S 4.4.
We train our GLCN for a maximum of 3000 epochs (training
iterations) using an ADAM algorithm \cite{Adam} with a learning rate of 0.005.
We stop training if the validation loss does not decrease for 100 consecutive
epochs, as suggested in \cite{kipf2016semi}.
All the network weights $\Theta$ are initialized using Glorot initialization~\cite{glorot2010understanding}.

\subsection{Comparison with state-of-the-art methods}

\textbf{Baselines.}  We first compare our GLCN model with GCN~\cite{kipf2016semi} which is the most related model with our GLCN.
We also compare our method against some other  graph neural network based semi-supervised learning approaches which contain i) two traditional graph based
semi-supervised learning methods including  Label Propagation (LP)~\cite{zhu2003semi}, Manifold Regularization (ManiReg)~\cite{belkin2006manifold},
and ii) three graph neural network methods including DeepWalk ~\cite{perozzi2014deepwalk}, Graph Convolutional Network (GCN)~\cite{kipf2016semi} and
Graph Attention Networks (GAT) ~\cite{velickovic2017graph}. 
The codes of these comparison methods were provided by authors and we can use them directly in our experiments.

\textbf{Results.} Table 1 summarizes the comparison results on three citation network benchmark datasets (Citeseer, Cora and Pubmed ~\cite{sen2008collective}).
Table 2 summarizes the comparison results on four widely used image datasets (CIFAR10~\cite{krizhevsky2009learning}, SVHN~\cite{netzer2011reading}, MNIST and Scene15 ~\cite{jiang2013label}). The best results are marked as bold in Table 1 and 2.
Overall, we can note that
(1) GLCN outperforms the baseline method GCN~\cite{kipf2016semi} on all datasets, especially on the four image datasets. This clearly demonstrates the higher predictive  accuracy  on semi-supervised classification of GLCN by incorporating graph learning architecture.
Comparing with GCN, the hidden layer presentations of graph nodes in GLCN become more discriminatively (as shown in Figure 4), which thus facilitates to semi-supervised learning results.
(2) GLCN performs better than recent graph network GAT ~\cite{velickovic2017graph}, which indicates the benefit of GLCN on graph data representation and learning.
(3) GLCN performs better than other graph based semi-supervised learning methods, such as LP~\cite{zhu2003semi}, ManiReg~\cite{belkin2006manifold}
and DeepWalk~\cite{perozzi2014deepwalk}, which further demonstrates the effectiveness of GLCN on conducting semi-supervised classification tasks on graph data.

\subsection{Parameter analysis}

In this section, we evaluate the performance of GLCN model with different settings of network parameter. 
We first investigate the influence of model depth of GLCN (number of convolutional layers) on semi-supervised classification results.
Figure 5 shows the performance of our GLCN method across different number of convolutional layers on MNIST dataset.
As a baseline, we also list the results of GCN model with the same setting.
One can note that
GLCN can obtain better performance with different number of layers, which indicates the insensitivity of the GLCN w.r.t. model depth.
Also, GLCN always performs better than GCN under different model depths, which further
demonstrates the benefit and better performance of GLCN comparing with the baseline method.
\begin{figure}[htpb]
\centering
\centering
\includegraphics[width=0.45\textwidth]{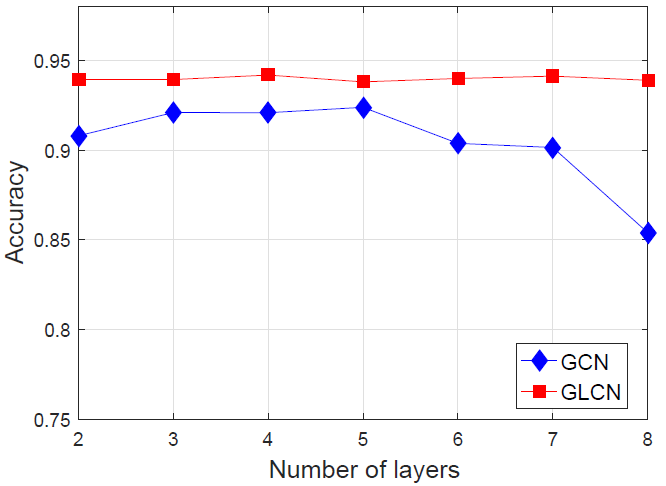}
  \caption{Results of GLCN across different convolutional layers on MNIST dataset. }\label{fig::lambda}
\end{figure}
\begin{table}[!htp]\small
\centering
\caption{Results of two-layer GLCN across different number of units in convolutional-layer on MNIST dataset.}\label{tb::patch}
\centering
\begin{tabular}{c|c|c|c|c|c}
\hline
\hline
GCN-Layers & 50 & 60 & 70 & 80 & 90 \\
\hline
GCN & 0.9041 & 0.9075 & 0.9080 & 0.9076 & 0.9070\\
\hline
GLCN & 0.9410 & 0.9396 & 0.9394 & 0.9410 & 0.9389\\
\hline
\hline
\end{tabular}
\end{table}
%

%
\begin{table}[!htp]\small
\centering
\caption{Results of GLCN with different settings of graph learning parameter $\lambda$ in loss (Eq.(14)) on MNIST and CIFAR10 datasets}\label{tb::patch}
\centering
\begin{tabular}{c|c|c|c|c|c|c}
\hline
\hline
Parameter $\lambda$& 0 & 1e-4 & 1e-3 & 1e-2 & 1e-1 & 1.0 \\
\hline
CIFAR10 &0.67 & 0.69 & 0.69 & 0.70 & 0.69 & 0.69\\
\hline
MNIST  &0.92 & 0.93 & 0.93 & 0.94 & 0.94 & 0.93\\
\hline
\hline
\end{tabular}
\end{table}
%


Then, we evaluate the performance of two-layer GLCN with different number of hidden units in convolutional layer.
Table 3 summarizes the performance of  GLCN with different number of hidden units on MNIST dataset.
We can note that
Both GCN and GLCN are generally insensitive  w.r.t. number of units in the hidden layer.

Finally, we investigate the influence of graph learning parameter $\lambda$ in our GLCN.
Table 4 shows the performance of GLCN with different parameter settings.
Note that, when $\lambda$ is set to 0, GLCN can also return a reasonable result.
Also, the graph learning regularization term in loss function will improve the graph learning and thus semi-supervised classification results.

\section{Conclusion and Future Works}

In this paper, we propose a novel Graph Learning-Convolutional Network (GLCN) for graph based semi-supervised learning problem.
GLCN integrates the proposed new graph learning operation and traditional graph convolution architecture together in a unified network, which
can learn an optimal graph structure that best serves GCN for semi-supervised learning problem.
%
%
Experimental results on seven benchmarks demonstrate that GLCN generally outperforms traditional fixed-graph CNNs on various semi-supervised learning tasks.
%
%
%

Note that, GLCN is not limited to deal with semi-supervised learning tasks.
In the future, we will adapt  GLCN on some more pattern recognition tasks, such as graph data classification, graph link prediction etc.
Also, we can explore GLCN method on some other computer vision tasks, such as visual object detection, image co-segmentation and visual saliency analysis. 
\bibliographystyle{ieee}
\bibliography{nmfgm1}

\end{document}